\documentclass[letterpaper, 10 pt, conference]{ieeeconf}  

\IEEEoverridecommandlockouts                              

\usepackage{amsmath} 
\usepackage{amssymb}  
\usepackage{amsfonts}
\usepackage[dvipsnames]{xcolor}
\usepackage{todonotes}
\usepackage[noadjust]{cite}
\usepackage{gensymb}
\usepackage{multirow}
\usepackage[table,xcdraw]{ colortbl}
\usepackage[ruled,linesnumbered,inoutnumbered]{algorithm2e}
\usepackage{graphicx}
\usepackage{caption}
\usepackage{subcaption}

\usepackage{lineno}

\usepackage{hyperref}
\hypersetup{
    colorlinks=true,
    linkcolor=magenta,
    filecolor=cyan,      
    urlcolor=blue,
    pdfpagemode=FullScreen,
}
\title{\LARGE \bf
Follow The Rules: Online Signal Temporal Logic Tree Search for Guided Imitation Learning in Stochastic Domains}

\author{Jasmine Jerry Aloor$^{*1}$, Jay Patrikar$^{*2}$,  Parv Kapoor$^{2}$,     Jean Oh$^{2}$ and Sebastian Scherer$^{2}$ 
\thanks{*equal contribution}
\thanks{This work was  supported by Army Futures Command.}
\thanks{$^{1}$Jasmine J. Aloor is with the Department of Aerospace Engineering, Indian Institute of Technology,
        Kharagpur, WB, India (work done while at Carnegie Mellon University)
        {\tt\small jasminejerry@iitkgp.ac.in}}
\thanks{$^{2}$Authors are with the Robotics Institute, Carnegie Mellon University, Pittsburgh, PA, USA. {\tt\small \{jpatrika, parvk, hyaejino, basti\}@andrew.cmu.edu}}%
}

\begin{document}

\maketitle
\thispagestyle{empty}
\pagestyle{empty}

\begin{abstract}
Seamlessly integrating rules in Learning-from-Demonstrations (LfD) policies is a critical requirement to enable the real-world deployment of AI agents.
Recently, Signal Temporal Logic (STL) has been shown to be an effective language for encoding rules as spatio-temporal constraints. 
This work uses Monte Carlo Tree Search (MCTS) as a means of integrating STL specification into a vanilla LfD policy to improve constraint satisfaction. We propose augmenting the MCTS heuristic with STL robustness values to bias the tree search towards branches with higher constraint satisfaction. While the domain-independent method can be applied to integrate STL rules online into any pre-trained LfD algorithm, we choose goal-conditioned Generative Adversarial Imitation Learning as the offline LfD policy. We apply the proposed method to the domain of planning trajectories for General Aviation aircraft around a non-towered airfield. Results using the simulator trained on real-world data showcase 60\% improved performance over baseline LfD methods that do not use STL heuristics.
\href{https://github.com/castacks/mcts-stl-planning}{[Code]}\footnote{ \href{https://github.com/castacks/mcts-stl-planning}{Codebase: https://github.com/castacks/mcts-stl-planning}}  
\href{https://youtu.be/fiFCwc57MQs}{[Video]}\footnote{ \href{https://youtu.be/fiFCwc57MQs}{Video: https://youtu.be/fiFCwc57MQs}}  
\end{abstract}

\section{INTRODUCTION}



\begin{figure}[ht]
    \centering
         \centering
        \includegraphics[width=\columnwidth]{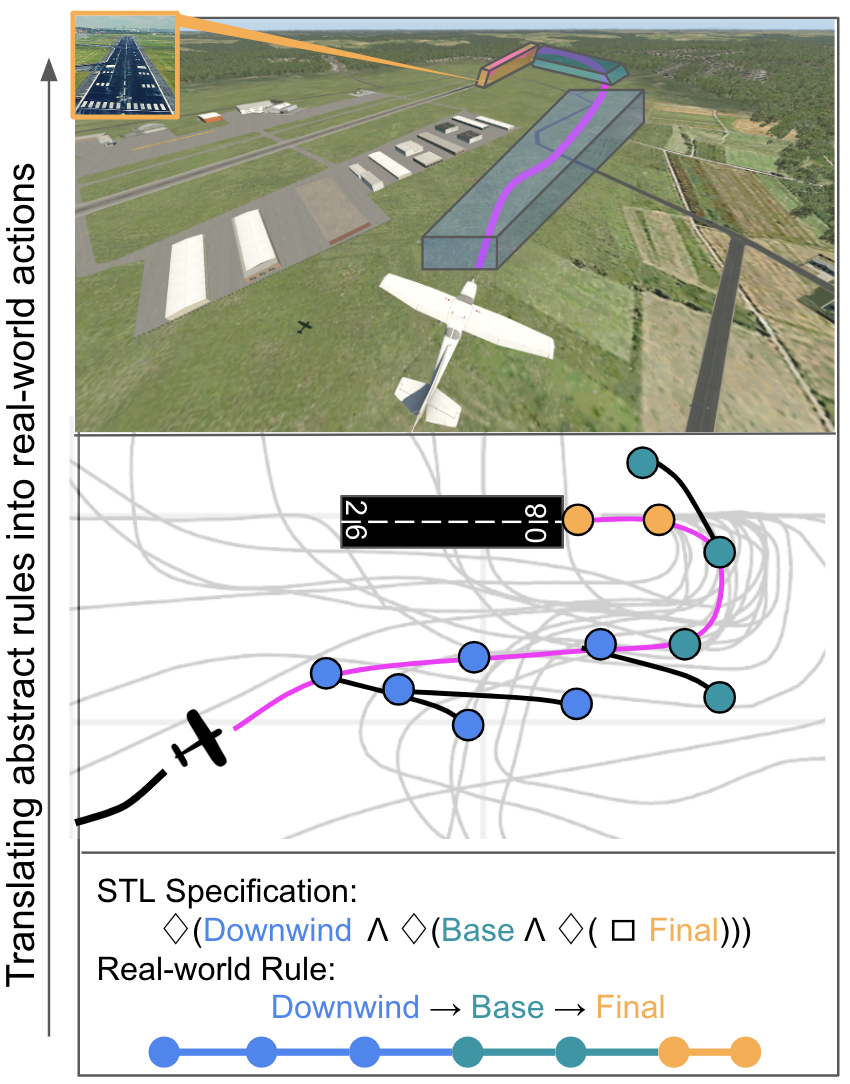}
        \caption{Figure shows the proposed approach in a prototypical scenario to plan for an aircraft landing in a non-towered airfield. The expert/pilot demonstrations (grey) are used offline to train an LfD policy; online, we use the Signal Temporal Logic specifications in the MCTS expansion to ensure rule compliance.}
        \vspace{-3.5mm}
    \label{fig:summary}
\end{figure}

 In many real-world domains, the ability to learn from multiple interactions with the environment is either prohibitively expensive or comes with safety concerns. Providing embodied AI agents with the ability to learn effectively from offline datasets of human demonstrations is thus critical in our pursuit to deploy them reliably in real-world domains.  Prior works have demonstrated limited success when Learning from Demonstrations (LfD) is treated as a purely supervised learning task~\cite{bagnell2015invitation}. Expert demonstrations are often noisy, incomplete, or both. Thus, pure LfD policies, like Behavior Cloning (BC), perform suboptimally when deployed in the real world. They also often fail to \textit{distill} the underlying rules and constraints that guide the behaviors of the expert agent. These issues are especially more pronounced when using LfD to train policies for safety critical systems when a goal is to be achieved while adhering to rules. We thus ask the question:
 \begin{center}
    \textit{ Can we encode rules as high-level task specifications to improve the online performance of the LfD policies?} 
 \end{center}
For continuous dynamical systems, there often exist spatio-temporal constraints that define the rules of operation of the system. However, traditional LfD methods do not provide a formal way to bias the system behavior for the satisfaction of rules. Moreover, the rules are often expressed in ambiguous natural language with partial specifications. \
Temporal logics such as Signal Temporal Logic (STL) provide a mathematically robust representation to encode such spatio-temporal constraints. They can be used to logically specify desired behavior translated from requirements expressed in natural language.  
STL is used to specify properties over real-valued dense time signals often generated by continuous dynamical systems \cite{maler2004monitoring}. 
Quantitative semantics associated with STL provides a real value called robustness which quantifies the degree of satisfaction or violation. 
This property enables the designer to encode domain-specific constraints and quantitatively measure their satisfaction with motion planning and control. 


In order to infuse the STL specifications to improve the base LfD, we propose using a variant of Monte Carlo Tree Search (MCTS) that uses an offline pre-trained network to generate simulated roll-outs. MCTS is a powerful heuristic search algorithm often deployed for long-horizon decision-making tasks \cite{AlphaGo}. 
MCTS, when used with a UCT heuristic, \cite{kocsis_bandit_2006} has properties like anytime convergence to the best action. This makes it well suited to be used as a long-horizon goal-directed planner within large state spaces with a time budget.

 \begin{figure*}[t]
    \centering
         \centering
        \includegraphics[width=\textwidth]{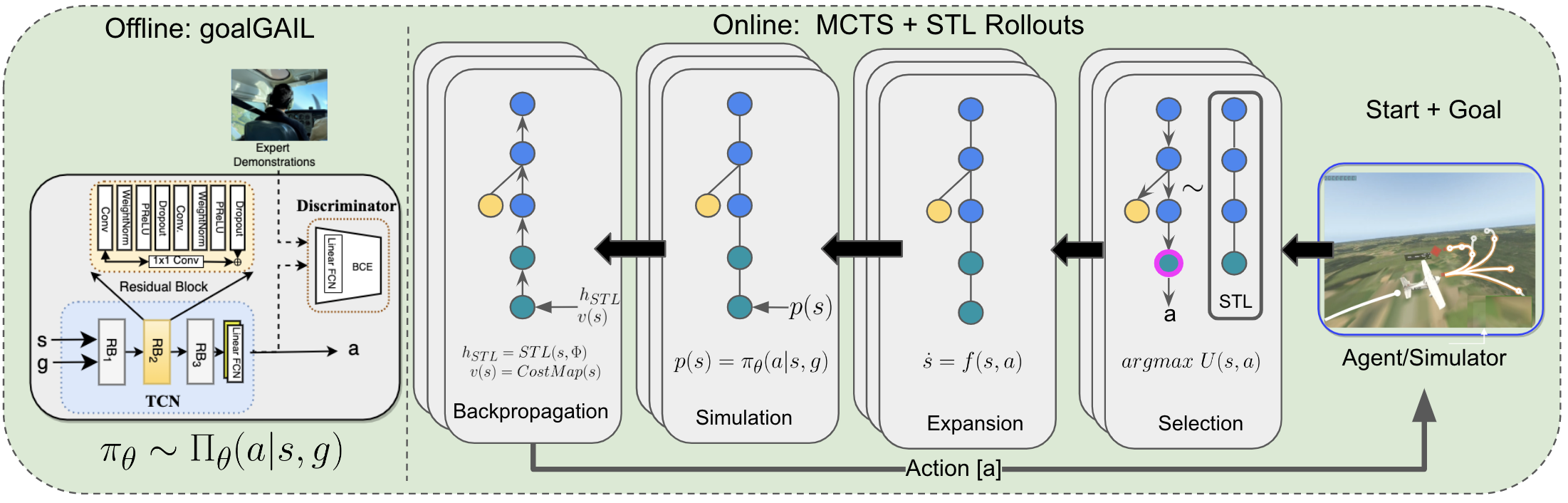}
        \caption{Overview of the approach: Offline, we train an LfD policy using datasets, which are used Online in a Monte-Carlo Tree Search (MCTS). The online expansion uses a modified heuristic that uses robustness values from Signal Temporal Logic (STL) specification to guide the search toward higher rule conformance.}
    \label{fig:mcts_overview}
\end{figure*}
In this work, we present a novel decision-making method that uses MCTS to build a tree structure that guides the offline pre-trained LfD policies online with STL specifications. We achieve this by biasing the MCTS heuristic sampling towards branches with higher STL satisfactions. The primary insight is that while the LfD policy decides on the low-level executions in line with the expert demonstrations, STL encourages the satisfaction of high-level objectives. This hierarchical approach provides a method to encode rules while allowing the agent to choose how to satisfy the constraints in a learning-enabled framework. The STL specifications thus provide a guide rail against the low-bandwidth noise in the expert demonstrations.   

In order to showcase the efficacy of the proposed algorithm, we use the prototypical case study of planning for a General Aviation (GA) aircraft operating at a non-towered airfield. GA aircraft operating under Visual Flight Rules (VFR) in non-towered airspaces are allowed to operate without a central authority while following some general rules established by the FAA. While not enforced, pilots are expected to adhere to these rules, but deviations arise due to a multitude of factors. Transponder data is available to observe and learn from this behavior, but it is often noisy as each pilot follows a variation of the rules. The proposed algorithm uses this data to train a behavior cloning policy and ensures rule compliance by augmenting the MCTS using STL specifications derived from the rules.   
Our contributions are as follows:
\begin{enumerate}
    \item We propose a novel method to incorporate high-level rules expressed in STL specifications in online MCTS simulation that augments any base pre-trained LfD policy.
    \item We showcase results on a challenging real-world problem that uses human demonstration data with experimental evaluations performed on a simulator and show improvements over the base policy.
\end{enumerate}





 The organization of the remainder of the paper is as follows: In Section \ref{sec:related_works}, we provide an overview of the previous approaches to solve sequential decision-making using imitation learning, MCTS, and STL. In Section  \ref{sec:method}, we introduce the approach and our algorithm. In Section \ref{sec:exper}, we provide implementation details for the GA problem statement. Section \ref{sec:eval} provides comparative results with established baselines. Section \ref{sec:conclusion} provides concluding remarks and outlines future work.

\section{Related Work}\label{sec:related_works}
Guiding LfD robot behavior using specifications expressed in formal logic has been previously explored in literature. Previous methods have leveraged LTL \cite{innes2020elaborating,wang2022temporal} or, more recently, STL \cite{leung2020back,kapoor2020model,yaghoubi2020worst,raman2014model} to encode the desired robot behavior to enable planning for autonomous systems. Most prior work focuses on offline backpropagating STL robustness along with imitation learning loss to improve the trained policy's constraint satisfaction. These proposed offline methods that learn from either a margin based on the lower bound of STL satisfaction \cite{cho2018learning}, reward functions \cite{puranic2021learning,puranic_learningSTL}, vector representation \cite{hashimoto2021stl2vec}, or risk metrics \cite{li2022learning}. 

While offline learning has led to improved STL satisfaction, there are no guarantees that the resulting controller will produce satisfying
trajectories \cite{leung2022semi} nor can it accommodate post hoc specification changes. To this end, methods that use constraints online in the form of Control Barrier Functions \cite{liu2021recurrent} and one-step state feedback \cite{yaghoubi2020worst} have been proposed, but neither uses expert demonstrations. \cite{igl2022symphony} takes an alternate approach to improving the efficiency and applicability of LfD-generated policies using beam search with goal generation as a hierarchical approach. The closest to our work is the recently proposed method \cite{leung2022semi} that uses STL and expert demonstrations to synthesize a trajectory-feedback controller. The offline component uses an LSTM-based controller whose parameters are modified on-the-fly. Our proposed method has no such requirements for using a particular architecture, nor does it require updating the base policy's parameters. 

While a majority of the algorithms showcase results on either grid-based worlds \cite{kalagarla2021model}, simple reach-avoid problems \cite{liu2021recurrent}, or deterministic settings \cite{kapoor2020model}, our work showcases results using real-world, noisy expert demonstrations. 

\section{Methodology}
\label{sec:method}
This section details the problem statement and the proposed framework.
\subsection{Preliminaries}
\label{subsec:prelim}
Given a continuous-space dynamic system of the form $\dot{{s}} = f(s,a)$, we define a discrete-time Markov Decision Process without rewards, (MDP \textbackslash R). Let $\mathcal{M} = (S,A,T, \rho_0, G)$ where S is the set of states $s \in S$, $ a \in A$ is the discrete set of actions or motion primitives that follow $f(\cdot), T : S \times A \Rightarrow S$ is the transition function, $\rho_0 \in S$ is an initial state distribution, and $G$ is the goal distribution. 

The task is to produce a policy $\pi(\theta)$ from a start location $s_0 \in \rho_0$ to goal location ${g} \in G$ that leads to a trajectory $\tau = (s_0,a_0,s_1,a_1, \dots, g)$. We also assume access to expert trajectories $\mathcal{D} =  \{(s^j_0,a^j_0,s^j_1,a^j_1, \dots, g^j)\}_{j=0}^D$ and high-level STL specification $\Phi$ that encodes any rules we expect the system to follow. An STL formula $\Phi$ can be built recursively from predicates using the following grammar
\begin{equation}
    \Phi := \top \ | \ \mu_c \ | \ \neg \Phi \ | \ \Phi \wedge \Psi \ | \ \diamondsuit_{[a,b]} \Phi \ | \ \Box_{[a,b]} \Phi \ | \ \Phi_1 \mathcal{U}_{[a,b]} \Phi_2
\end{equation}
where $\Phi_1, \Phi_2$ are STL formulas, $\top$ is the Boolean True, $\mu_c$ is a predicate of the form $\mu(s) > c$, $\neg$ and
$\wedge$ the Boolean negation and AND operators, respectively, and 0 $\le$ a $\le$ b $ < \infty$ denote time intervals. The temporal operators $\diamondsuit, \Box $ and $\mathcal{U}$
are called ``eventually", ``always", and ``until" respectively. The quantitative semantics of a formula with respect to a signal $\vec{x_t}$ can be used to compute robustness values \cite{donze2013efficient} for the specifications used in our application.

\subsection{Framework}
The overall framework is shown in Fig. \ref{fig:mcts_overview} and Algorithm \ref{alg:plan}. We first train an LfD policy by formulating the problem as finding a distribution of future actions  ${\hat{a}_t}$ conditioned on the past trajectories ${{s}}_{t-t_{obs}:t}$ and the goal ${{g}}$ where $t_{obs}$ is the observation time horizon.

\begin{equation}
    \hat{\pi}_\theta \sim \Pi_\theta( {\hat{a}}_{t} \mid {{s}}_{t-t_{obs}:t}, g)
    \label{actions_future}
\end{equation}  

The MCTS (see Algorithm \ref{alg:MCTS}) uses the policy $\hat{\pi}_\theta$ to generate simulations. 
Each simulation starts from the root state and iteratively selects moves that maximize the STL-modified UCT heuristic.

For each state transition, we maintain a directed edge in the tree with an action value $Q(s,a)$,  prior probability $P(s, a)$, STL heuristic $H(s,a)$, and a visit count $N(s, a)$. The total heuristic value is calculated as 
\begin{equation}
   U(s,a) =  Q(s,a') + \frac{c_1P(s,a')\sqrt{N(s)}}{1+N(s,a')} + c_2H(s,a')
\end{equation}
where $c_1$ and $c_2$ are the hyperparameters controlling the degree
of exploration and STL heuristic's weight. Starting with the initial state $s_{(0)}$, at each time step, we calculate the action to take, which maximizes $U (s, a)$ (Line 12).  If the next state already exists in the tree, we continue our simulation, else a new node is created in our tree, and we initialize its
$P (s, \cdot)$ = $\vec p_\theta (s)$ from
our policy $\hat{\pi}_\theta$ (Line 8). The expected reward $v = v_\theta  (s)$ can be provided by the user as a learned value function or as a cost-map (Line 9). The heuristic $h_{STL}$  is calculated using the STL specification (Line 14). We  initialize $Q(s, a)$, $H(s, a)$  and $N (s, a)$
to 0 for all $a$. We then propagate the cost $v$ and the STL heuristic $h_{STL}$ back up the MCTS tree, updating all the $Q(s, a)$ and $H(s, a)$ values seen
during the simulation, and start again from the root. 


After running forward simulations of the MCTS, the $N (s, a)$ values provide a good approximation for the optimal stochastic process from each state. Hence, the action we take is randomly sampled from a distribution of actions with
probability proportional to ${N (s, a) }$. We expand the search space until we exhaust the planning budget time $planHorizon$. After each action is selected, the MCTS tree is reinitialized from the actual trajectory followed by the agent. 
 The planner is terminated when the goal is reached, or the maximum number of steps is reached, whichever comes first. 
 \vspace{-3mm}
\RestyleAlgo{ruled}

\SetKwComment{Comment}{/* }{ */}
\begin{algorithm}
\caption{Plan ($\theta$,$\Phi$)}
\label{alg:plan}

\SetAlgoLined
$s \gets Sample(\rho_0)$\\
$g \gets Sample(G)$\\
\LinesNumbered
\While{$s \neq g \ \mathbf{and}$ not maxSteps}{
\While{$timeElapsed \le planHorizon$}{
    $N(\cdot) \gets MCTS(s_{0:t},g,\theta,\Phi,0)$
}
$a \gets choice_{a'} (N(s,a'))$\\
$s \gets T(s,a)$
}
\end{algorithm}
\begin{algorithm}
\caption{MCTS ($s_{0:t}$,g,$\theta$,$\Phi$,$h_{STL}$)}
\label{alg:MCTS}

\If{s $\in$ G }{\Return s == g, $h_{STL}$}

\eIf {s $\notin$ Tree}{
$Tree \gets Tree \cup s$\\
$Q(s,a) \gets 0$\\
$N(s,a) \gets 0$\\
$P(s,a) \gets \hat{\pi}_\theta(s)$\\
$v(s) \gets CostMap(s)$\\

\Return $v(s), h_{STL}$\\
}{
$a \gets argmax_{a'}\Big[Q(s,a') + \frac{c_1P(s,a')\sqrt{N(s)}}{1+N(s,a')} + c_2H(s,a')\Big]$\\
$s' \gets T(s,a)$\\
$h_{STL} \gets STL(s' + s_{0:t},\Phi)$\\
$v,h_{STL} \gets MCTS(s' + s_{0:t},g,\theta,\Phi,h_{STL})$\\
$N(s,a) \gets N(s,a) + 1$\\
$Q(s,a) \gets \frac{ N(s,a)Q(s,a) + v}{1+N(s,a)}$\\
$H(s,a) \gets \frac{N(s,a)*H(s,a) + h_{STL}}{1+N(s,a)}$\\
\Return $v , h_{STL}$
}
\end{algorithm}
\vspace{-7mm}
\section{Experiments}\label{sec:exper}

In our experiments, we tackle the case of planning for a general aviation aircraft at a non-towered airport. This section provides the necessary background, problem definition, and implementation details. 
\subsection{Background}

In an environment without a centralized controlling authority, such as near a non-towered airspace, the FAA establishes rectangular airport traffic patterns. These patterns facilitate the smooth flow of aircraft entering and leaving the airspace,  similar to roadways for automobiles.
However, different pilots follow distinct paths \cite{patrikar_moon_ghosh_oh_scherer_2021}, leading to variations in the expert demonstration trajectories.  This variation arises due to the unique combinations of aircraft type, weather, pilots' experience, visual observations, and the non-strict regulations for following the pattern. A single fixed set of waypoints
to follow would not suffice to cover all possibilities for the aircraft. With such a broad set of rules, a range of possible trajectories can be followed to land an aircraft safely.

To illustrate with an example, the high-level goal of landing on a runway through traffic patterns can usually be divided into three stages \cite{FAA}. The three stages can be represented by three individual regions to be occupied by the aircraft in a particular order (See Fig. \ref{fig:goalrep}). The first stage is termed the downwind leg, which involves flying in the opposite direction to the intended landing runway at 1000 ft above ground level. Once the pilot is at a 45$\degree$ angle from the approach end of the runway, the second stage begins, in which a perpendicular turn is made to the base leg until it reaches the runway approach's extended center line. Once the turn is complete, the third and final stage begins, which is a descending path to the point of touchdown. We provide a mechanism to encode such traffic patterns and utilize LfD to implement the sequence of actions that can land an aircraft from a given start to the runway, just like a human pilot.
%


\vspace{-1mm}
\subsection{Problem Definition}
Consider a fixed-wing aircraft at time $t$, let ${s}_{t} = (x_t,y_t,z_t,\chi_t) \in \mathcal{R}^3 \times SO(2)$ denote the position and heading of the agent. We also define the system $\dot{{s}} = f({s},{a})$ \eqref{eq:kinematics} as:
\begin{subequations}
\begin{align}
 \dot{x}&=v_{2D}\cos \chi\\ 
  \dot{y}&=v_{2D} \sin \chi\\ 
 \dot{z}&=v_h\\
 \dot{\chi} &= \frac{g_e \tan \phi }{v_{2D}}\\
 v &= \sqrt{v_{2D}^2 + v_h^2}
\end{align}\label{eq:kinematics}
\end{subequations}
where $v$ is the aircraft's inertial speed, $v_{2D}$ is the speed in the 2-D plane, $\phi$ is the roll-angle, and $v_h$ is vertical speed. Finally, $g_e$ is the acceleration due to gravity. We ignore the effects of wind.

\begin{figure}
     \centering
    \includegraphics[width=0.9\columnwidth]{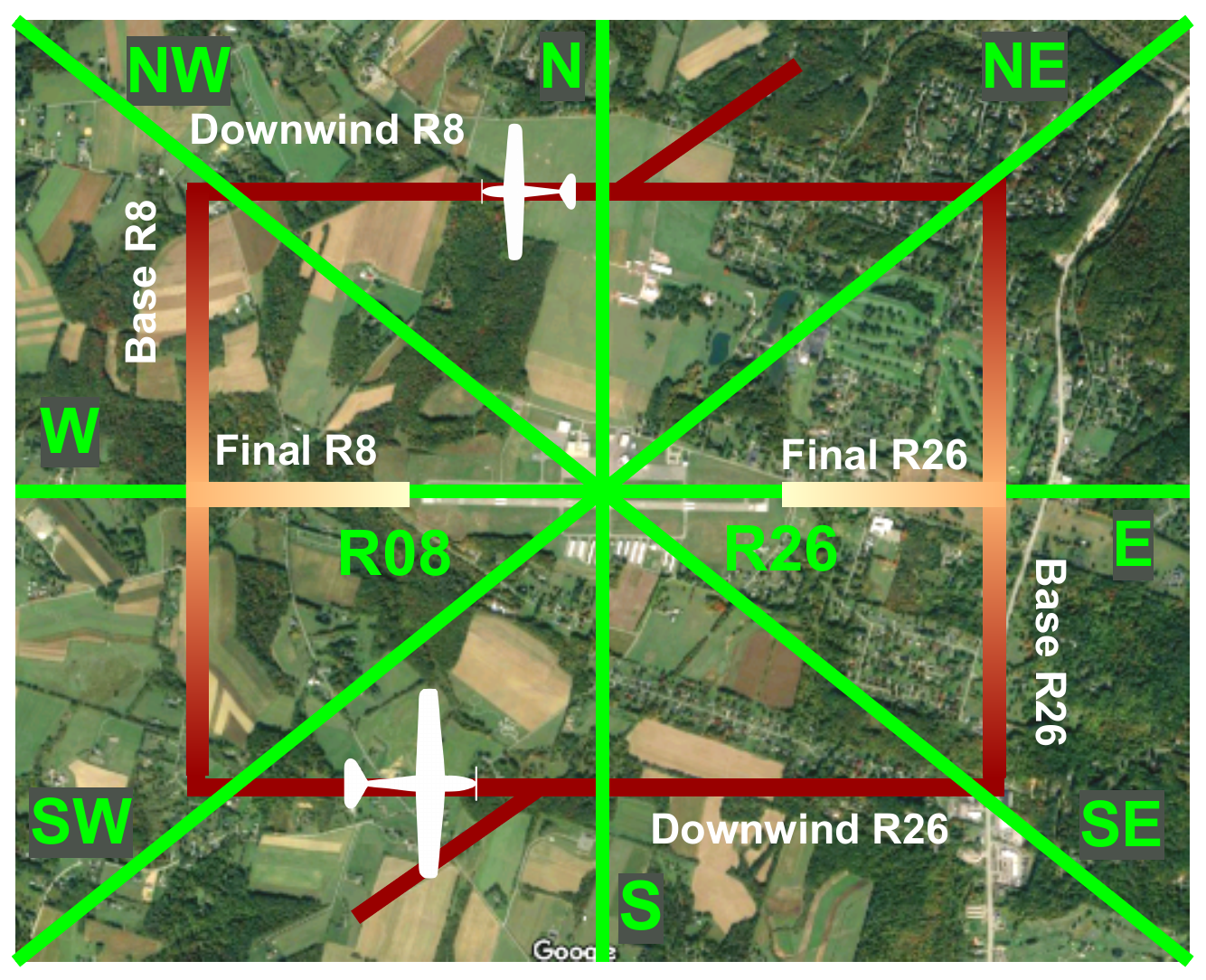}    

      \caption{Goal representation is in the form of a one-hot vector where each region is the respective goal element in the goal vector $G $. The maroon rectangle shows airport traffic patterns.} \label{fig:goalrep}
\end{figure}

The action space ${A} =  \{(v^j,v_h^j,\phi^j ) \}_{j=0}^A$ is a fixed library of 30 motion primitives that discretize each of the control inputs. 
We use the inertial velocities ($v$)  70 and 90 knots, the vertical velocities ($v_h$)  +500 ft/min and -500 ft/min, and the bank angle ($\phi$) discretization such that $\chi$ changes by $ 45\degree$ and $90\degree$ heading over the chosen time-horizon of 20 sec.
The goal distribution $G$ is a one-hot vector representation of the final goal of a particular agent as the eight cardinal directions along with two runway ends, as shown in Fig. \ref{fig:goalrep}. The set $G$ is represented as 
$G = \{ N, NE, E, SE, S, SW, W, R08, R26\}$ with each element representing the final region the airplane is desired to reach as shown in Fig. \ref{fig:goalrep}. For simplicity, we also set the start states to one of these regions, i.e.\ $\rho_0 = G$. 

\subsection{Implementation Details}
The implementation details are split between online and offline components. 

\subsubsection{{Dataset}} We use the \textit{TrajAir}\footnote{\href{http://theairlab.org/trajair/}{http://theairlab.org/trajair/}} dataset, which provides recorded trajectories of aircraft  operating at the Pittsburgh-Butler Regional Airport (ICAO:KBTP) \cite{patrikar_moon_ghosh_oh_scherer_2021}. 
The dataset contains 111 days of transponder data, processed to obtain the local $x,y$, and $z$ coordinates of aircraft at every second. This is used to extract expert demonstrations from pilots as they navigate the un-towered airspace. The dataset trajectories are smoothed using a B-spline (basis-spline) approximate representation of order 2. The trajectories are not biased toward satisfying the STL rules as there are variations in landing patterns exhibited by the dataset.
 
\subsubsection{{Offline LfD Policy  Details}}\label{sec:model}
The LfD policy takes as input the past trajectories of the agent to predict its possible action distribution. While the method can use any LfD policy, we use a goal-conditioned generative adversarial imitation learning (GoalGAIL) method \cite{ding2019goal}. The GoalGAIL is modified to use Temporal Convolutional Layers (TCNs) to process the sequential trajectory data. TCN layers encode a trajectory's spatio-temporal information into a latent vector without losing the underlying data's temporal (causal) relations \cite{TCN}. We use TCNs as an alternative to using LSTMs \cite{aircraft_LSTM} for encoding the trajectories. 

We break the trajectories in a scene into sequences of length $t_{obs}+t_{pred}$ where $t_{obs} = 11 sec$ and $t_{pred} = 20 sec$. In a given scene, the raw trajectory in absolute coordinates of the agent is encoded using the TCN layers as  ${h}_{obs}$. The agent's goal ${g \sim G}$ is encoded through an MLP layer, $\varphi_1$, and is concatenated with the encoded trajectory vector. Equations \ref{eq:TCN} show the encoding sequence. 
\begin{subequations}
\begin{align}
{h}_{obs} &= TCN_{obs} ({{s}}_{1:t_{obs}}) \\ 
{h}_{g} &= \varphi_1({g})\\
{h}_{enc} &= {h}_{obs} \oplus  {h}_{g}\\
 {\hat{s}}_{t_{obs}:t_{obs}+t_{pred}} &= \varphi_2({h}_{enc})
\end{align}\label{eq:TCN}
\end{subequations}
The $\mathcal{L}_{act}$ measures how close the predicted trajectory is to the expert trajectory using a mean squared error (MSE) loss.
\begin{equation}
    \mathcal{L}_{act} = MSE({s}_{t_{obs}:t_{obs}+t_{pred}},{\hat{s}}_{t_{obs}:t_{obs}+t_{pred}})
\end{equation}

\begin{table*}[]
\centering
\begin{tabular}{|
>{\columncolor[HTML]{FFFFFF}}c |
>{\columncolor[HTML]{EFEFEF}}c 
>{\columncolor[HTML]{FFFFFF}}c 
>{\columncolor[HTML]{EFEFEF}}c 
>{\columncolor[HTML]{FFFFFF}}c |
>{\columncolor[HTML]{EFEFEF}}c 
>{\columncolor[HTML]{FFFFFF}}c 
>{\columncolor[HTML]{EFEFEF}}c 
>{\columncolor[HTML]{FFFFFF}}c |
>{\columncolor[HTML]{EFEFEF}}c |}
\hline
\cellcolor[HTML]{EFEFEF}{\color[HTML]{000000} }                                     & \multicolumn{4}{c|}{\cellcolor[HTML]{EFEFEF}{\color[HTML]{000000} \textbf{Takeoff Specification $\Phi_T$}}}                                                                                                                                                                                                                                 & \multicolumn{4}{c|}{\cellcolor[HTML]{EFEFEF}{\color[HTML]{000000} \textbf{Landing Specification $\Phi_L$}}}                                                                                                                                                                                                                                 & \cellcolor[HTML]{EFEFEF}{\color[HTML]{000000} }                                 \\ \cline{2-9}
\multirow{-2}{*}{\cellcolor[HTML]{EFEFEF}{\color[HTML]{000000} \textbf{Algorithm}}} & \multicolumn{1}{c|}{\cellcolor[HTML]{EFEFEF}{\color[HTML]{000000} N}}                  & \multicolumn{1}{c|}{\cellcolor[HTML]{FFFFFF}{\color[HTML]{000000} S}}                  & \multicolumn{1}{c|}{\cellcolor[HTML]{EFEFEF}{\color[HTML]{000000} E}}                  & {\color[HTML]{000000} W}                  & \multicolumn{1}{c|}{\cellcolor[HTML]{EFEFEF}{\color[HTML]{000000} N}}                  & \multicolumn{1}{c|}{\cellcolor[HTML]{FFFFFF}{\color[HTML]{000000} S}}                  & \multicolumn{1}{c|}{\cellcolor[HTML]{EFEFEF}{\color[HTML]{000000} E}}                  & {\color[HTML]{000000} W}                  & \multirow{-2}{*}{\cellcolor[HTML]{EFEFEF}{\color[HTML]{000000} \textbf{Total}}} \\ \hline
{\color[HTML]{000000} BC + MCTS}                                                    & \multicolumn{1}{c|}{\cellcolor[HTML]{EFEFEF}{\color[HTML]{000000} 0.2 / 0.2}}          & \multicolumn{1}{c|}{\cellcolor[HTML]{FFFFFF}{\color[HTML]{000000} 0.0 / 0.1}}          & \multicolumn{1}{c|}{\cellcolor[HTML]{EFEFEF}{\color[HTML]{000000} 0.3 / 0.1}}          & {\color[HTML]{000000} 0.3 / 0.1}          & \multicolumn{1}{c|}{\cellcolor[HTML]{EFEFEF}{\color[HTML]{000000} 0.0 / 0.0}}          & \multicolumn{1}{c|}{\cellcolor[HTML]{FFFFFF}{\color[HTML]{000000} 0.1 / 0.4}}          & \multicolumn{1}{c|}{\cellcolor[HTML]{EFEFEF}{\color[HTML]{000000} 0.1 / 0.4}}          & {\color[HTML]{000000} 0.3 / 0.4}          & {\color[HTML]{000000} 0.1 / 0.2}                                                \\ \hline
{\color[HTML]{000000} GoalGAIL + MCTS}                                              & \multicolumn{1}{c|}{\cellcolor[HTML]{EFEFEF}{\color[HTML]{000000} 0.0 / 0.3}}          & \multicolumn{1}{c|}{\cellcolor[HTML]{FFFFFF}{\color[HTML]{000000} 0.1 / 0.4}}          & \multicolumn{1}{c|}{\cellcolor[HTML]{EFEFEF}{\color[HTML]{000000} 0.0 / 0.3}}          & {\color[HTML]{000000} 0.0 / 0.4}          & \multicolumn{1}{c|}{\cellcolor[HTML]{EFEFEF}{\color[HTML]{000000} 0.3 / 0.6}}          & \multicolumn{1}{c|}{\cellcolor[HTML]{FFFFFF}{\color[HTML]{000000} 0.1 / 0.2}}          & \multicolumn{1}{c|}{\cellcolor[HTML]{EFEFEF}{\color[HTML]{000000} 0.3 / 0.7}}          & {\color[HTML]{000000} 0.1 / 0.5}          & {\color[HTML]{000000} 0.1 / 0.4}                                                \\ \hline
{\color[HTML]{000000} BC +  MCTS + STL}                                             & \multicolumn{1}{c|}{\cellcolor[HTML]{EFEFEF}{\color[HTML]{000000} 1.0 / 0.9}}          & \multicolumn{1}{c|}{\cellcolor[HTML]{FFFFFF}{\color[HTML]{000000} 1.0 / 0.9}}          & \multicolumn{1}{c|}{\cellcolor[HTML]{EFEFEF}{\color[HTML]{000000} 0.6 / 0.3}}          & {\color[HTML]{000000} 0.7 / 0.4}          & \multicolumn{1}{c|}{\cellcolor[HTML]{EFEFEF}{\color[HTML]{000000} 0.2/ 0.5}}           & \multicolumn{1}{c|}{\cellcolor[HTML]{FFFFFF}{\color[HTML]{000000} 0.2 / 0.5}}          & \multicolumn{1}{c|}{\cellcolor[HTML]{EFEFEF}{\color[HTML]{000000} 0.2 / 0.5}}          & {\color[HTML]{000000} 0.2 / 0.5}          & {\color[HTML]{000000} 0.5 / 0.6}                                                \\ \hline
{\color[HTML]{000000} GoalGAIL + MCTS+STL}                                          & \multicolumn{1}{c|}{\cellcolor[HTML]{EFEFEF}{\color[HTML]{000000} \textbf{1.0 / 0.9}}} & \multicolumn{1}{c|}{\cellcolor[HTML]{FFFFFF}{\color[HTML]{000000} \textbf{1.0 / 0.9}}} & \multicolumn{1}{c|}{\cellcolor[HTML]{EFEFEF}{\color[HTML]{000000} \textbf{0.8 / 0.7}}} & {\color[HTML]{000000} \textbf{1.0 / 0.9}} & \multicolumn{1}{c|}{\cellcolor[HTML]{EFEFEF}{\color[HTML]{000000} \textbf{0.7 / 0.9}}} & \multicolumn{1}{c|}{\cellcolor[HTML]{FFFFFF}{\color[HTML]{000000} \textbf{0.8 / 0.9}}} & \multicolumn{1}{c|}{\cellcolor[HTML]{EFEFEF}{\color[HTML]{000000} \textbf{0.3 / 0.8}}} & {\color[HTML]{000000} \textbf{0.5 / 0.9}} & {\color[HTML]{000000} \textbf{0.7 / 0.8}}                                       \\ \hline
\end{tabular}
\caption{Table shows the quantitative results with randomly sampled start and goal states for two LfD policies with ablation studies with the STL heuristic. Results show the Success Rate $\uparrow$ / STL Score $\uparrow$ for two vanilla LfD policies and their ablations with the STL heuristic. Results show both Landing $X \Rightarrow R$ and Takeoff $R \Rightarrow X$ scenarios for each of the cardinal directions $X$.    }
\label{table:results}
\end{table*}
A discriminator \( D_{\psi} \) is trained to distinguish expert transitions $(s, a, g)   \sim \tau_{\text {expert }},  E({{s}}_{1:t_{obs}}) $, from policy transitions \( (s, a, g) \sim \tau_{\text {policy }},  \hat{s}_{t_{obs}:t_{obs}+t_{pred}} \). 
The discriminator is trained to minimize,
\begin{equation}
\begin{aligned}
    \mathcal{L}_{goalG A I L}\left(D_{\psi}, \right) 
    =& \mathbb{E}_{(s, a, g) \sim \mathrm{policy}}\left[\log D_{\psi}(a, s, g)\right]+ \\ 
    & \mathbb{E}_{(s, a, g) \sim \mathrm{expert}}\left[\log \left(1-D_{\psi}(a, s, g)\right)\right]  
    \end{aligned}
\end{equation}

The combination of these two loss functions is used to train the model.
\begin{equation}
    \mathcal{L}_{total} = \mathcal{L}_{act} + \mathcal{L}_{goalG A I L}
\end{equation}
In order to convert $\hat{s}$ to $\hat{a}$, we match the generated trajectories from the control inputs in the library $A$ using a weighted L2 Euclidean error distance over $(x, y, z)$ points on the trajectory. For training, we use the AdamW optimizer with a learning rate of $3e-3$.

\begin{figure*}[h!]
    \centering
    \includegraphics[width=\textwidth]{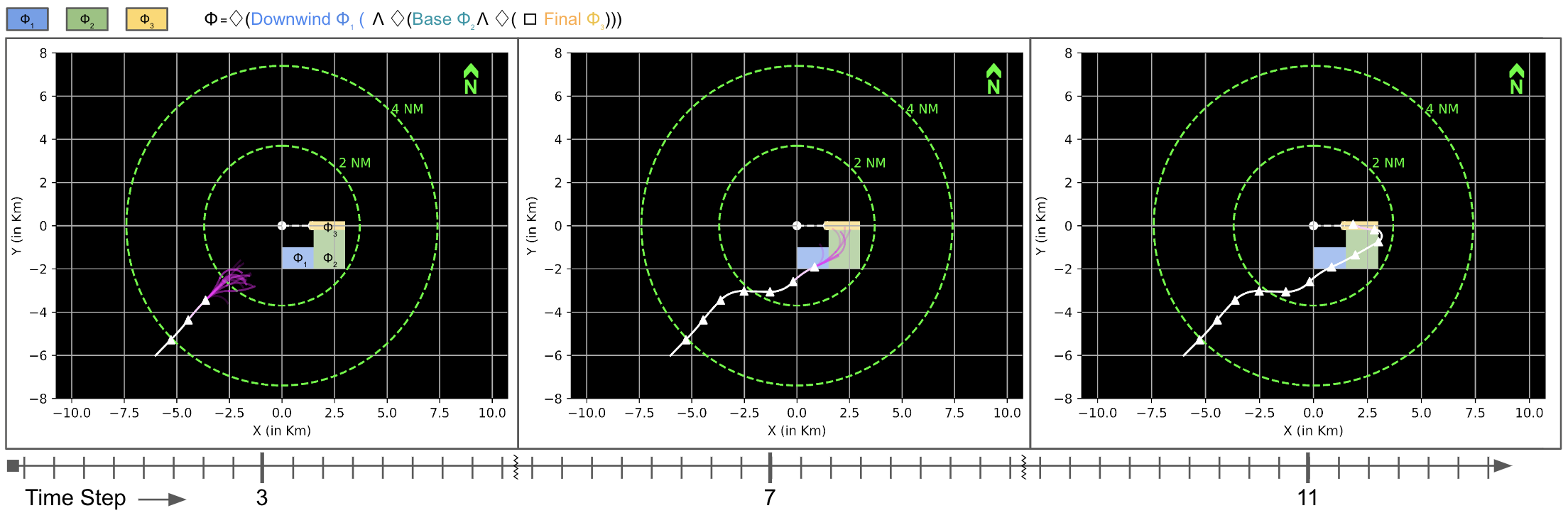}
    \caption{Figure shows a qualitative example from one of the cases where the aircraft starts from the South-West and needs to land at one of the runways (R26). The specifications \textcolor{cyan}{$\Phi_1$}, \textcolor{SpringGreen}{$\Phi_2$} and \textcolor{Goldenrod}{$\Phi_3$} are shown as rectangles. White marked lines show the aircraft trajectory, and the \textcolor{magenta}{magenta} shows the MCTS tree. The runway threshold for R08 (+x-axis) is at the center. }
    \label{fig:res}
\end{figure*}
\subsubsection{{Signal Temporal Logic Specifications}}
We evaluate the performance of our agent based on reaching the goal while adhering to the airport traffic pattern. The goal objective, as well as traffic pattern compliance, are both encoded using STL specifications. 
We use the three stages for the landing pattern as defined in (IV-A). $\Phi_1$, $\Phi_2$, and $\Phi_3$ represent the STL formulas encoding occupancy of regions corresponding to the downwind, base, and final stages, respectively. 
The landing STL specification becomes:
\begin{equation}
    \Phi_L = \diamondsuit( \Phi_1 \ \wedge \  (\diamondsuit( \Phi_2\  \wedge  \ (\diamondsuit  \square  \Phi_3  ))))
\end{equation}
$\Diamond( \Phi$) can be interpreted as ``Eventually" being in a region represented by $\Phi$. The nested $\Diamond$ operators encode a sequential visit of regions represented by $\Phi_1$, $\Phi_2$, and $\Phi_3$. Similarly, the takeoff STL specification is defined based on the goal regions reached by the aircraft. By defining reaching a goal region  $g \sim G$  by an STL formula $\Phi_4$, we get the takeoff specification:
\begin{equation}
    \Phi_T = \diamondsuit( \Phi_4 )\ 
\end{equation}

The robustness values of the STL specifications are evaluated on the state trajectory traces generated by the search tree. The first element of the traces is the tree root node, and the last element of the trace is the node whose value is currently computed.
\subsubsection{Online Monte Carlo Tree Search}
The MCTS is implemented as a recursive function where each iteration ends with a new leaf that corresponds to an action in the trajectory library. The implementation uses a normalized costmap $v(s)$ that is built by counting the frequencies of the aircraft in the TrajAir dataset at particular states $s$.   
\section{Evaluations}

\label{sec:eval}

Evaluation of the proposed approach is done using a custom simulator that follows the dynamics defined in Eq. \ref{eq:kinematics}. The network implementations are done on PyTorch. For calculating STL robustness values, we formulate our specifications using the \verb+rtamt+ \cite{nivckovic2020rtamt} package, an online monitoring library.
To showcase real-time online evaluations, simulations are performed on an Intel NUC computer with Intel® Core™ i7-8559U CPU @ 2.70GHz × 8. The complete implementation details and parameter details are in the open-sourced code-base\footnote{ \href{https://github.com/castacks/mcts-stl-planning}{Codebase: https://github.com/castacks/mcts-stl-planning}} and the associated Readme.
\subsection{Qualitative results}
Figure \ref{fig:res} shows an example scenario where the aircraft starts from the South-West and is tasked with landing at R26 while following the standard FAA traffic pattern. The rules of the traffic pattern are encoded as STL specifications. The white-marked line shows the aircraft's position at each step. At every step, the MCTS replans, and the resulting tree is shown in magenta. The STL sub-specifications are shown as rectangles. As can be seen, the aircraft manages to reach the runway while satisfying the specifications. The size of the search tree is a function of available $planHorizon$.  
\subsection{Comparative results}
In order to perform quantitative studies, we compare the performance of the proposed algorithm with vanilla LfD policies. We uniformly sample 100 start-goal pairs randomly from $\rho_0, G$. $G$ is truncated to ${N,S,E,W,R}$ where $R$ represents both R08 and R26 to condense the results. We then provide these to Algorithm \ref{alg:plan}, which plans for the agent. Based on the selected start and goal pair, a relevant STL specification is chosen. Each episode ends when the agent reaches the goal or if the maximum steps are exceeded. In addition to the goalGAIL policy, we also train a pure Behavior Cloning (BC) policy. The BC policy uses a TCN to encode the history and outputs an action without a goal vector or a discriminator. Comparisons were carried out with both these policies integrated into the MCTS framework with ablation on the STL heuristic to show the impact of the STL specification.
 
We define our evaluation metrics as follows:
\begin{itemize}
    \item Success Rate: Fraction of episodes that were successful in reaching their goal locations.
    \item STL Score: Average of the normalized fraction of the STL robustness value satisfied over all the episodes. A higher value indicates better satisfaction. 
\end{itemize}

Table  \ref{table:results}  shows  the quantitative results. Our proposed algorithm performs significantly better than the baselines for all start-goal pairs. We get an almost perfect success rate in the aircraft takeoff scenarios. The aircraft landing cases are more challenging due to following the specific landing patterns when incoming from different sides of the runway, which is reflected in the success rates shown. Additionally, we observe the baseline BC performs similarly to GAIL on the success metric but not on the STL robustness. This indicates that while BC policies are comparable in reaching the goals, the transient performance of GoalGAIL is better. Incorporating STL improves the robustness values for both LfD policies.  

\section{CONCLUSIONS AND FUTURE WORK}
\label{sec:conclusion}
In this work, we present a novel method that improves the online robustness of offline pre-trained LfD policies using MCTS to fuse STL specifications. To the best of the authors' knowledge, this is the first method that combines high-level STL specifications with low-level LfD policies through MCTS. Our experimental evaluation targets the real-world problem of autonomous aircraft planning and exhibits the feasibility of our techniques for similar challenging decision-making problems. We show our method outperforms vanilla LfD methods on a number of successful missions for multiple complex objectives using real-world data. 

While the results are promising, future work involves producing theoretical guarantees on the satisfaction of the STL specifications. The current work biases the search tree towards satisfying STL constraints but provides no guarantees. Another line of work is to validate the algorithm on a high-fidelity simulator and real-world flight tests. 





\section{Acknowledgment}
The authors would like to thank Joao Dantas for his help with the costmaps.
\addtolength{\textheight}{-10cm}   

\bibliographystyle{IEEEtran}
\bibliography{./IEEEfull,root}
\end{document}